\documentclass[letterpaper, 10 pt, conference]{ieeeconf}

\usepackage{xcolor}

\IEEEoverridecommandlockouts                              % This command is only needed if 
\title{\LARGE \bf
Correct-by-Construction Advanced Driver Assistance Systems\\ based on a Cognitive Architecture
}

\author{Francisco Eiras$^{1}$, Morteza Lahijanian$^{2}$, and Marta Kwiatkowska$^{3}$% <-this % stops a space
\thanks{This work was partially supported by EPSRC Mobile Autonomy Program Grant EP/M019918/1. FiveAI provided a travel grant to support the presentation of this work.}
\thanks{$^{1}$Francisco Eiras was a student at the Dept. of Computer Science, University of Oxford, UK, when this work was developed and is now with FiveAI
        {\tt\small francisco.eiras@five.ai}}%
\thanks{$^{2}$Morteza Lahijanian is with the Dept. of the Ann and H.J. Smead Aerospace Engineering Sciences, University of Colorado Boulder
        {\tt\small morteza.lahijanian@colorado.edu}}%
\thanks{$^{3}$Marta Kwiatkowska is with the Dept. of Computer Science, University of Oxford, UK
        {\tt\small marta.kwiatkowska@cs.ox.ac.uk}}%
}

\usepackage{amssymb}
\usepackage{amsmath}
\usepackage{graphicx}
\usepackage{amsthm}
\usepackage{mathtools}
\usepackage{float}
\usepackage{lscape}
\usepackage{url}
\usepackage{multirow}
\usepackage{cite}
\usepackage{hyperref}

\theoremstyle{definition}

\theoremstyle{definition}
\newtheorem{defi}{Definition}
\newtheorem{problem}{Problem}

\newcommand{\bigmid}{\:\mid\:}
\newcommand{\cond}{\:\mid\:}

\newcommand{\near}{\text{near}}
\newcommand{\far}{\text{far}}
\newcommand{\car}{\text{car}}
\newcommand{\follow}{\text{follow}}
\newcommand{\hw}{\text{hw}}

\newcommand{\act}{\mathrm{a}}

\graphicspath{ {figures/} }
\usepackage{caption}
\usepackage{subcaption}

\begin{document}

\maketitle
\thispagestyle{empty}
\pagestyle{empty}

%%%%%%%%%%%%%%%%%%%%%%%%%%%%%%%%%%%%%%%%%%%%%%%%%%%%%%%%%%%%%%%%%%%%%%%%%%%%%%%%
\begin{abstract}
Research into safety in autonomous and semi-autonomous vehicles has, so far, largely been focused on testing and validation through simulation. Due to the fact that failure of these autonomous systems is potentially life-endangering, formal methods arise as a complementary approach. This paper studies the application of formal methods to the verification of a human driver model built using the cognitive architecture ACT-R, and to the design of correct-by-construction Advanced Driver Assistance Systems (ADAS).  
% The approach is based on probabilistic modeling, and 
The novelty lies in the integration of ACT-R in the formal analysis and an abstraction technique that enables finite representation of a large dimensional, continuous system in the form of a Markov process.  The situation considered is a multi-lane highway driving scenario and the interactions that arise.  The efficacy of the method is illustrated in two case studies with various driving conditions.
% The results conclusively show that the ADAS designed improves safety and time efficiency in this scenario. 
\end{abstract}

%%%%%%%%%%%%%%%%%%%%%%%%%%%%%%%%%%%%%%%%%%%%%%%%%%%%%%%%%%%%%%%%%%%%%%%%%%%%%%%%
\section{Introduction}

% \ml{Need a more exciting opening sentence!, e.g., Autonomous cars are still the dream of the future.  Despite the general belief, today's technology has not yet been able to support a fully autonomous car that works in every city and driving condition.  Semi-autonomous cars, on the other hand, are here ...}
% \fge{We can't open with a sentence like that, I work for an AV company! Maybe something a little more moderate that drives the point across that humans are not good drivers}
Humans do not have a good track record on the road. Road accidents kill 1.24 million people every year and over $90\%$ of all crashes are mainly attributed to errors of human drivers \cite{crashes}. While full self-driving technology is not yet available at scale, in an attempt to reduce these numbers, several car manufacturers have introduced semi-autonomous features in the form of Advanced Driver Assistance Systems (ADAS). Examples include Tesla's \textit{Autopilot} and Ford's \textit{Co-Pilot 360}. However, ensuring safety for semi-autonomous vehicles remains a major challenge with roots in the lack of coherent understanding of the human-ADAS interaction. 
% In this study, we focus on this issue.
% \ml{end with the general problem domain..., e.g., Ensure safety for semi-autonomous vehicles, however, remains a major challenge with roots in lack of coherent understanding of human-ADAS interaction. In this study, we focus on this challenge.}

% While ADAS are more practical for drivers and would appear beneficial in terms of safety, there are two concerns regarding the way these systems are designed: they are built by humans and the presence of software mistakes in the implementation is almost unavoidable, particularly due to the complexity of the systems; and existing ADAS lack the capability of understanding the human cognitive process; hence, they are unable to predict the actions drivers will take and adjust or suggest safe actions accordingly.

% \subsection{Related Work}

% One approach to validate the safety of these systems would be to test drive the ADAS in real world situations. \ml{maybe this topic sentence instead: ``Existing methods to ... rely on testing and simulation''}
Existing methods to validate the safety of semi-autonomous systems rely on testing and simulation. Using real data to take statistically significant conclusions, however, is infeasible due to the time it takes to collect a sufficiently large amount of data \cite{driving-to-safety}. Several approaches are based on modeling and simulating the semi-autonomous vehicle, as proposed in \cite{sim1, sim2, sim3, sim4}. Despite this, it is imperative to recognize the shortcomings of simulation in safety evaluation of complex driver assistance systems which could have life-endangering impact \cite{challenges1, challenges2}.

% Several works have introduced formal verification techniques as a way to evaluate control systems in semi-autonomous and autonomous vehicles \cite{games, fv2}. \ml{Stronger emphasis on how good formal methods are... formal methods to rescue..., e.g., A promising direction is to employ formal verification techniques, which are based on rigorous mathematical reasoning to ... , as proposed by several recent works [cite].}
A promising direction is to employ formal verification techniques, which are based on rigorous mathematical reasoning, to obtain strong guarantees about the ADAS, as proposed by several recent works \cite{games, nilsson, ver_human_behavior}. In \cite{nilsson}, Nilsson \textit{et al.} synthesized a provably-correct module for adaptive cruise control from specifications given in temporal logic. However, these studies ignore 
% \ml{``overlook'' is too weak... lack of reasonable human model is a major issue...}\fge{but do we want to go further and possibly risk insulting one of the reviewers if we are talking about their work?} 
the human driver behavior variability presented by a driver model, which can lead to controllers that perform poorly in corner cases. On the other hand, \cite{ver_human_behavior} applies model checking techniques to the verification of data-driven models of human driver behavior, yet it does not explicitly model the human cognitive process nor does it leverage this analysis as a way to bootstrap safety in the form of an ADAS.
% \ml{... also their human model lacks cognition process}

% \subsection{Summary}

\begin{figure}
\centering
\includegraphics[width=0.47\textwidth]{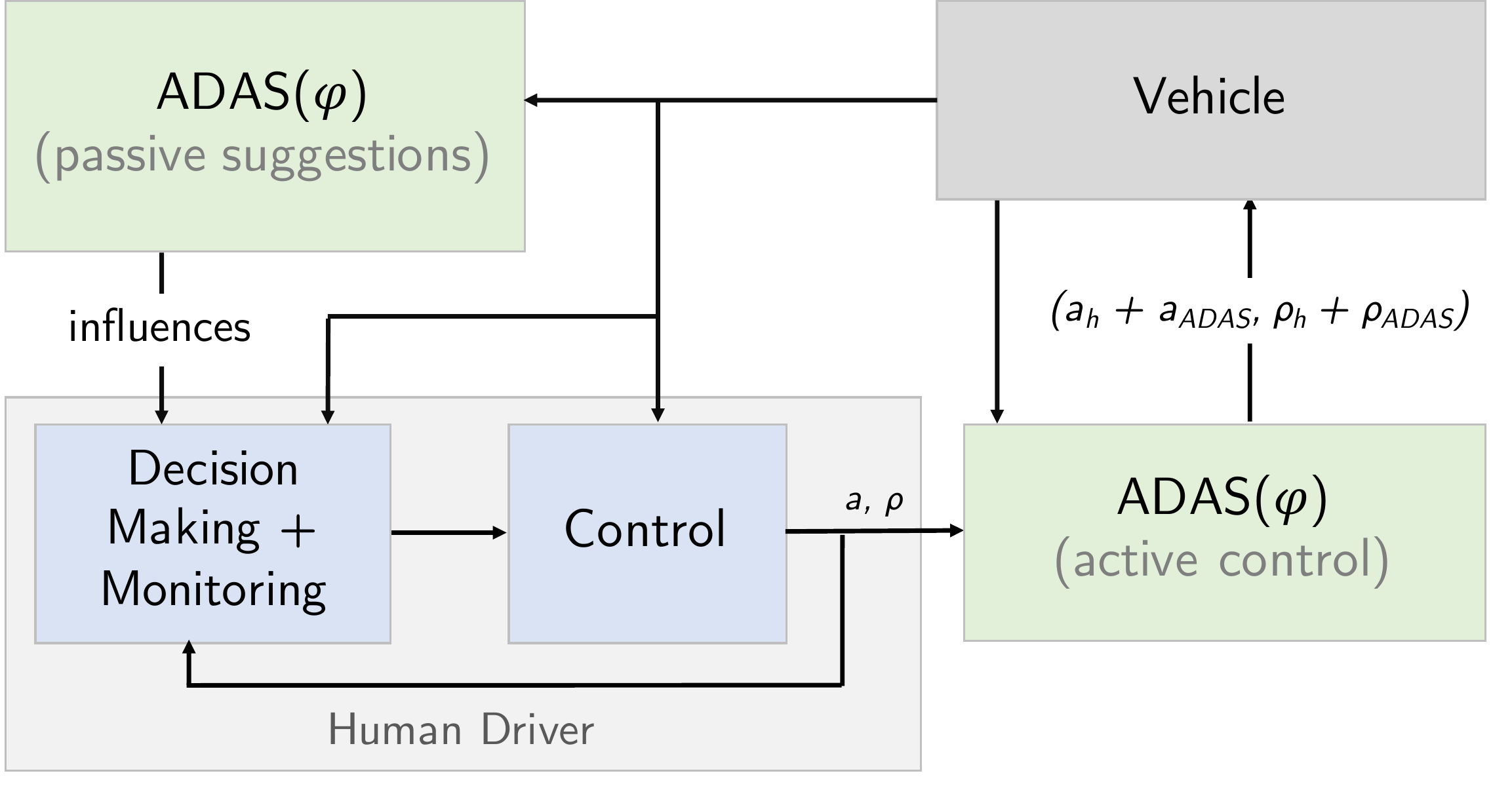}
\caption{
Overview of the ADAS design with passive and active interventions for a specification $\varphi$.
% Overview of the final system and the intervention of the ADAS (synthesized based on a PCTL path formula, $\varphi$, and its maximization or minimization, $\bowtie \in \{\max, \min\}$) in the sequential (solid arrows) and informational (dashed arrows) flows.
}
\label{fig:final_system}
\end{figure}

% \ml{start with the goal of the study, then narrow it down to what we do in this paper..., e.g., the overarching goal of this study is to provide safety guarantee in autonomous and semi-autonomous vehicles through integration of human cognition and rational with formal methods.  As a first step in this direction, this paper focuses on ...}
The overarching goal of this study is to provide safety guarantees in semi-autonomous vehicles through the integration of human cognition with formal methods. As a first step in this direction, this paper focuses on giving guarantees at the design level of the ADAS. 
% \ml{Specifically, it employs the human cognitive model known as Adaptive Control of Thought—Rational (ACT-R) [cite] and builds on ...}
Specifically, it employs the cognitive architecture known as Adaptive Control of Thought-Rational (ACT-R):
a framework for specifying computational behavioral models of human cognitive performance, embodying both the abilities (e.g. memory storage and recall, perception or motor action) and constraints (e.g. memory decay and limited motor performance) of humans \cite{actr_1, actr_2, salvucci_0, salvucci_1, actr_3, actr_4}.
The work builds on the human driver model in a multi-lane highway driving scenario presented in \cite{salvucci_1}. It also expands upon \cite{games} by applying verification techniques to an efficient abstraction of the model and extends it to allow the intervention of a provably-correct synthesized ADAS based on specifications given as temporal logic formulas.

% \subsection{Contribution}

The main contribution of this paper is threefold: first, it studies the verification of a human driver model built in a cognitive architecture through efficient model abstraction techniques. Second, it builds upon the model of human driving behavior as a way to bootstrap the desired properties in the ADAS using formal methods. Third, it introduces a flexible framework in terms of specifications which allows for different guarantees to be obtained depending on the choices made by the ADAS designer. Other contributions of this work include case studies based on specific properties and an open source implementation of the framework. To the best of our knowledge, this is the first framework that brings formal reasoning to the design of semi-autonomous vehicle solutions by taking into account the cognitive process of the human.
% \ml{To the best of our knowledge, this is the first framework that allows formal reasoning of human-autonomy system that takes into account the cognitive process of the human.}
% \ml{Isn't your MATLAB-prism tool a contribution?}
% \fge{Other contributions include case studies for .... and the implementation of the framework is available}

%%%%%%%%%%%%%%%%%%%%%%%%%%%%%
%%%%%%%%%%%%%%%%%%%%%%%%%%%%
\section{Problem Formulation}
\label{sec:problem-formulation}
% \ml{give a high-level description of the scenario..., e.g., We consider a semi-autonomous driving scenario in a highway, where ...}
% \fge{to simplify the problem, we assume the motions of other vehicles are known; if there's data, then we can use it to better predict the behaviors of others}

We consider the driving scenario studied in \cite{salvucci_1}, where a vehicle, called the \textit{ego-vehicle}, is in an interaction with a lead vehicle in a  multi-lane highway.  We are interested in designing a correct-by-construction ADAS system for the ego-vehicle.

% It should be noted that the two-vehicle scenario is non-limiting as traffic in highways tends to be sparse and it is acceptable to reason over each of the other vehicles encountered separately \fge{is this confusing? I meant there are generally few vehicles and it might be fine to just consider them pairwise with us}.

% We assume the motion of the other vehicle to be deterministic and, to simplify the problem, known to the ego-vehicle. This is a reasonable assumption due to the predictability of highway driving and the possible improvements that can be introduced by using data. The two-vehicle assumption is also non-limiting, as traffic in highways tends to be sparse and it is acceptable to reason over each of the other vehicles encountered separately \fge{is this confusing? I meant there are generally few vehicles and it might be fine to just consider them pairwise with us}.

\subsection{Vehicle Model}
\label{sec:kinematics}

% We assume the vehicle to be represented by a discrete kinematic bicycle model as described in \cite{vehicle_dynamics}. A state of the vehicle is defined as $\mathbf{x} = (x, y, v, \psi)$, where $x, y$ are the coordinates of the centre of mass in an inertial frame $\mathcal{F}~\in~\mathbb{R}^2$, $v$ is the speed and $\psi$ is the inertial heading of the vehicle; the control inputs are a tuple $\mathbf{u} = (\rho, a)$ where $\rho$ is the angle of the current velocity of the center of mass with respect to the longitudinal axis of the car (steering angle) and $a$ is the acceleration of the vehicle. The discrete difference equations that represent the kinematic bicycle model are:

We consider the ego-vehicle kinematics are described by
% assume the vehicle to be represented by a discrete kinematic bicycle model as described in \cite{vehicle_dynamics}. A state of the vehicle is defined as $\mathbf{x} = (x, y, v, \psi)$, where $x, y$ are the coordinates of the centre of mass in an inertial frame $\mathcal{F}~\in~\mathbb{R}^2$, $v$ is the speed and $\psi$ is the inertial heading of the vehicle; the control inputs are a tuple $\mathbf{u} = (\rho, a)$ where $\rho$ is the angle of the current velocity of the center of mass with respect to the longitudinal axis of the car (steering angle) and $a$ is the acceleration of the vehicle. The discrete difference equations that represent the kinematic bicycle model are:
\begin{equation}
\label{eq:car-kinematics}
    \begin{split}
        \Delta x &= v\cos(\psi + \rho) \Delta t, \quad \quad
        \Delta y = v\sin(\psi + \rho) \Delta t,\\
        \Delta v &= a \Delta t, \hspace{23.5mm} 
        \Delta \psi = \frac{2v}{l}\sin(\rho)\Delta t,
    \end{split}
\end{equation}
where $x$ and $y$ are the coordinates of the vehicle's center of mass, $v$ is the speed, and $\psi$ is the heading angle of the vehicle.  The control inputs are  
% $\rho$, the angle of the current velocity of the center of mass with respect to the longitudinal axis of the car (steering angle) 
steering angle $\rho$
and 
% $a$, the acceleration of the vehicle
acceleration $a$.
Finally, $l$ is the length of the vehicle, and $\Delta t$ is the time duration between two iterations of the model.

We assume that the motion of the lead vehicle is predictable.  This simplifying assumption, even though not realistic in large scale, is reasonable for small road segments due to the predictability of highway driving and the possible improvements that can be introduced by using data \cite{highway_pred}.

\subsection{Integrated Human Driver Model in ACT-R}

The ego-vehicle is driven by a human, whose behavior is represented in ACT-R.  ACT-R is 
a framework for specifying computational behavioral models of human cognitive performance \cite{actr_1, actr_2, salvucci_0, salvucci_1, actr_3, actr_4}.
It embodies two crucial cognitive aspects of humans: the abilities (e.g., memory storage, perception, and motor action) and the constraints (e.g. memory decay and limited motor performance).
ACT-R can be generally described as two distinct layers: a perceptual-motor layer and a cognitive layer.  The perceptual-motor layer corresponds to the interface of the cognition with the environment, being comprised of modules such as vision and motor actions. The cognitive layer is focused on memory, which can be divided into two different categories: declarative (consisting of factual knowledge and goals - e.g., \textit{``The maximum driving speed in a typical US highway is 65 mph"} or \textit{``Try to get to point B"}) and procedural (consisting of rules/procedures - e.g., \textit{``If the lead vehicle is going slowly, attempt an overtake"}) \cite{actr_1}.

\begin{figure}
\centering
\includegraphics[width=0.48\textwidth]{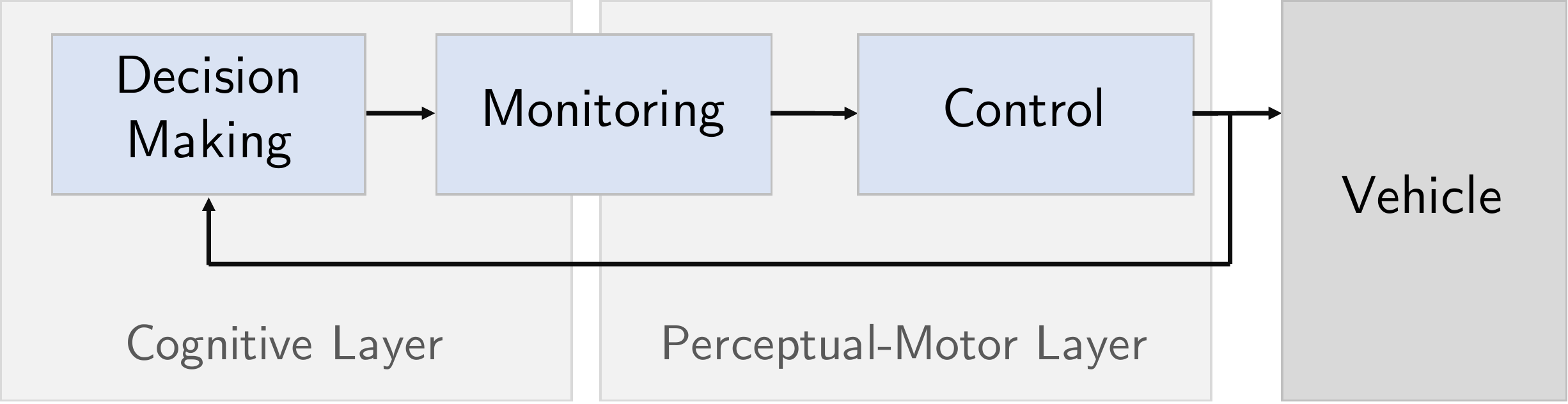}
\caption{Schematic overview of the interaction of the \textit{control}, \textit{monitoring} and \textit{decision making} modules of the integrated human driver model in ACT-R.}
\label{fig:salvucci_actr}
\end{figure}

Particularly, we focus on the model proposed by \cite{salvucci_1}, which is an improved version of the model from \cite{salvucci_0} based on advances in ACT-R and real world data. It describes how a human controls a vehicle and performs an action (e.g. lane change), in the presence of other vehicles. The model, shown schematically in Fig.~\ref{fig:salvucci_actr}, consists of three distinct modules interacting in a sequential way:
\begin{itemize}
    \item \textit{control}, which manages both the lower level perception cues and the physical manipulation of the vehicle;
    \item \textit{monitoring}, which maintains situational awareness through the awareness of the position of other vehicles around the ego-vehicle; and 
    \item \textit{decision making}, which uses the information gathered in the monitoring and control stage to determine the tactical decision to be taken (whether or not a lane change should happen). 
\end{itemize}

A full description of the model and the governing equations for vehicle control generations are provided in Sec. \ref{sec:verification}.

\subsection{ADAS Design}
\label{sec:ADAS}

We consider an ADAS that corresponds, first and foremost, to determining the possible available interventions to the system at each point in time.  The actions considered must be realistic in nature; otherwise, the obtained assistance system would prove to be incompetent in a real-world scenario. Fig.~\ref{fig:final_system} summarizes the interventions we consider for the ADAS, which are divided into two types: \textit{passive suggestions} and \textit{active control}.

In passive suggestions, it is assumed that the assistance system cannot change the decision making directly (as it is a human cognitive process), but it can influence it to a certain degree through suggestions \cite{driver_behavior}.  Hence, the control inputs to the vehicle are directly provided by the human (ACT-R), i.e., $a = a_h$ and $\rho=\rho_h$, where subscript $h$ corresponds to the human, and the ADAS can only provide suggestions that can lead to safe and correct behaviors.

In active control, ADAS can have incremental control-based interventions at the level of acceleration and steering, i.e., $a = a_h + a_\text{ADAS}$ and $\rho = \rho_h + \rho_\text{ADAS}$, where the ADAS variables are constrained to ensure incremental interventions.
The full details of the action availability, constraints, and intervention of the ADAS are presented in Sec.~\ref{sec:synthesis}.

% The design of a correct-by-construction ADAS corresponds, first and foremost, to determining the possible available interventions to the system at each point in time. The actions considered must be realistic in nature, otherwise the obtained assistance system would prove to be incompetent in a real-world scenario. Fig.~\ref{fig:final_system} summarizes the interventions we consider for the ADAS, which are divided into two types:
% \begin{itemize}
%     \item \textit{passive suggestions}: it is assumed that the assistance system can not change the decision making directly (as it is a human cognitive process), but it can influence it to a certain degree through suggestion \cite{driver_behavior};
%     \item \textit{active control}: incremental control-based interventions at the level of acceleration and steering.
% \end{itemize}

% The full details regarding the actions available and intervention of the ADAS are presented in Sec.~\ref{sec:synthesis}.

% \ml{Add a new subsection called ADAS Design (or something like it) where you explain ADAS (passive and active).  It is important to describe what you mean by ADAS so that it will be clear what it means in the problem.}
% \fge{What does it mean to have passive suggestions vs active control (at most two paragraphs)}

% \subsection{DTMCs, MDPs and Temporal Logic}
\subsection{Specification Language}
% \ml{it may be better to move the defs of DTMC/MDP/Adv to either a new section called preliminaries or the approach section.   Just define PCTL here... call the subsection ``Behavior Specification''}

% \ml{give some context... To formally express the behavioral properties of interest of the semi-autonomous car, such as safety and reachability, we use \textit{Probabilistic Computation Logic Tree} (PCTL) [cite].  A PCTL formula combines boolean and temporal operators with probabilistic reasoning, constituting a rich specification language.}

To formally express the behavioral properties of interest for the semi-autonomous system, we use \textit{Probabilistic Computation Tree Logic} (PCTL) \cite{bk08}. A PCTL formula combines boolean and temporal operators with probabilistic reasoning, constituting a rich specification language.

\begin{defi}[PCTL Syntax]
  A \textit{PCTL formula} $\Phi$ over a set of atomic propositions $AP$ can be formed according to the following grammar:
  \begin{align*}
    \Phi &:= \text{\texttt{true}} \bigmid o \bigmid \Phi \wedge \Phi \bigmid \Phi \vee \Phi \bigmid \neg \Phi \bigmid \mathbb{P}_{\sim p} (\varphi) \\
    \varphi &:= \bigcirc \:\Phi \bigmid \Phi_1 \mathcal{U}^{\leq k}\, \Phi_2 \bigmid \Phi_1 \mathcal{U} \Phi_2 \bigmid \Diamond \:\Phi
  \end{align*}
  where $o \in AP$ is an atomic proposition, 
  $\wedge$ (``and''), $\vee$ (``or''), and $\neg$ (``negation'') are boolean operators, 
  and $\bigcirc$ (``next''), $\mathcal{U}^{\leq k}$ (``bounded until'') with $k\in \mathbb{N}$, $\mathcal{U}$ (``until''), and $\Diamond$ (``eventually'') are temporal operators.  
  $\mathbb{P}$ is the  probabilistic operator, and $\sim p$ is a probability bound. The formulae $\Phi$ and $\varphi$ are called state and path formalas, respectively.
\end{defi}

% where $a \in AP$ \ml{$a$ is already used for acceleration. use something else.}, $\varphi$ is a path formula, $\sim p$ is a probability bound, and $\mathbb{P}_{\sim p}$ is the  probabilistic operator.

% A \textit{PCTL path formulae} $\varphi$ over a set of atomic propositions $AP$ can formed according to the following grammar:
% \begin{equation}
% 	\varphi ::= \bigcirc \:\Phi \bigmid \Phi_1 \mathcal{U}^{\leq k}\, \Phi_2 \bigmid \Phi_1 \mathcal{U} \Phi_2 \bigmid \Diamond \:\Phi
% \end{equation}
% where $a \in AP$, $k \in \mathbb{N}$, $\Phi_i$ are PCTL state formulas, $\bigcirc$ is the next operator, $\mathcal{U}$ is the until operator, $\mathcal{U}^{\leq k}$ is the bounded-until operator and $\Diamond$ is the eventually operator.
% \end{defi}

In this work, the atomic propositions represent boolean facts about the driving scenario. Through them, we can express properties of interest using PCTL, e.g., ``The probability that eventually the distance to the nearest car becomes less than $d_\text{safe}$ is less than 0.001'' can be expressed as $ \mathbb{P}_{< 0.001}\big( \Diamond (\parallel \mathbf{x} - \mathbf{x}_\text\near \parallel < d_\text{safe}) \big)$.

% \ml{describe how atomic propositions relate to the scenario with examples... We define atomic propositions to represent a fact about the autonomous driving scenario that can be either true or false.  Therefore, we can express properties of interest using PCTL, e.g., ``The probability that eventually the distance to the nearest car becomes less than $d_\text{safe}$ is less than 0.001'' can be expressed as $ \mathbb{P}_{< 0.001}\big( \Diamond (\parallel \mathbf{x} - \mathbf{x}_\text\near \parallel < d_\text{safe}) \big)$.} 

% For a given path $\pi = s_0s_1s_2...$ in a discrete-time Markov chain $\mathcal{M} = (S, \mathbf{P}, p_i, AP, L)$ (generalising trivially to a Markov decision process):
% \begin{itemize}
% 	\item $\pi \models \bigcirc\Phi$ if, and only if, $s_1 \models \Phi$.
% 	\item $\pi \models \Phi_1\mathcal{U}^{\leq k}\, \Phi_2$ if, and only if:
% 	 \begin{equation}
% 	 	\exists i \leq k \text{ s.t. } (s_i \models \Phi_2) \wedge (\forall j < i: s_j \models \Phi_1)
% 	\end{equation}
% 	\item $\pi \models \Phi_1\mathcal{U}\Phi_2$ if, and only if:
% 	 \begin{equation}
% 	 	\exists i \geq 0 \text{ s.t. } (s_i \models \Phi_2) \wedge (\forall j < i: s_j \models \Phi_1)
% 	\end{equation}
% \end{itemize}

\subsection{Problem Statement}
\label{sec:problem-statement}

Given a vehicle, whose motion is described by \eqref{eq:car-kinematics},  
% the continuous integrated driver model of highway behavior in ACT-R, 
and a human driver represented by ACT-R, 
a set of initial conditions defined as a scenario $\mathcal{S}$, 
% and a specification as a pair $(\varphi, \bowtie)$, where $\varphi$ is a PCTL path formula and $\bowtie~\in~\{\max, \min\}$,
and a PCTL formula $\varphi$, we are interested in the following two problems:

% \begin{itemize}
%   \item[\textit{1.}] (\textit{verification})\textit{:} \hspace{1mm} 
%   compute the probability that $\varphi$ is satisfied in $\mathcal{S}$, i.e, $\mathbb{P}_{=?}^{\mathcal{S}} (\varphi)$, and
%   \item[\textit{2.}] (\textit{synthesis})\textit{:} \hspace{1mm}
%   design an ADAS that optimizes the probability of satisfying $\varphi$ in $\mathcal{S}$, i.e., $\mathbb{P}_{\bowtie=?}^{\mathcal{S}} (\varphi)$ with $\bowtie~\in~\{\max, \min\}$.
% \end{itemize}

\begin{problem}[verification]
\label{problem:1}
Compute the probability that the human-vehicle system satisfies $\varphi$ in $\mathcal{S}$, i.e, $\mathbb{P}^{\mathcal{S}} (\varphi)$.
\end{problem}

\begin{problem}[synthesis]
\label{problem:2}
Design an ADAS that optimizes the probability of satisfying $\varphi$ by the human-vehicle-ADAS system in $\mathcal{S}$, i.e., $\mathbb{P}_{\bowtie}^{\mathcal{S}} (\varphi)$ with $\bowtie~\in~\{\max, \min\}$.
\end{problem}

This is a flexible problem representation under which the specification $\varphi$ comes from the designer of the ADAS. It should be noted that the two-vehicle scenario considered in this study is non-limiting as traffic in highways tends to be sparse, allowing to reason over each of the  vehicles separately as in \cite{salvucci_1}.  In addition, the proposed solution to Problems \ref{problem:1} and \ref{problem:2} is general and can be easily extended to more vehicles.

%%%%%%%%%%%%%%%%%%%%%%%%%%%%%%%%%
%%%%%%%%%%%%%%%%%%%%%%%%%%%%%%%%%
\section{Preliminaries}
In this study, we employ Markov models as the abstractions for the driving scenarios.

\begin{defi}[Markov Chain (MC)]
\label{def:MC}
A MC is a tuple $\mathcal{M} = (S, \mathbf{P}, s_0, AP, \iota)$, where $S$ is a finite set of states, $\mathbf{P}: S \times S \to [0,1]$ is a transition probability function, $s_0 \in S$ is the initial state, $AP$ is a set of atomic propositions, and $\iota: S \to 2^{AP}$ is a labelling function.
% \begin{itemize}
% 	\item $S$ is a countable, non-empty set of states
% 	\item $\mathbf{P}: S \times S \to [0,1]$ is a transition probability function such that for all states $s \in S: \sum_{s' \in S} \mathbf{P}(s,s') = 1$,
% 	\item $p_i: S \to [0,1]$ is the initial state distribution, such that $\sum_{s \in S} p_i(s) = 1$
% 	\item $AP$ is a set of atomic propositions, and 
% 	\item $L: S \to 2^{AP}$ is a labelling function. 
% \end{itemize}
% $\mathcal{M}$ is denoted \textit{finite} if $S$ and $AP$ are finite sets.
\end{defi}

\begin{defi}[Markov Decision Process (MDP)]
\label{def:MDP}
An MDP is a tuple $\mathcal{M} = (S, Act, \mathbf{P}, s_0, AP, \iota)$, 
where $S$, $s_0$, $AP$, and $\iota$ are as in Definition \ref{def:MC}, $Act$ is a finite set of actions, and $\mathbf{P}: S \times Act \times S \to [0,1]$ is a transition probability function.
% where $S$ is a countable, non-empty set of states; $Act$ is a set of actions; $\mathbf{P}: S \times Act \times S \to [0,1]$ is a transition probability function; $s_0$ is the initial state; $AP$ is a set of atomic propositions, and $\iota: S \to 2^{AP}$ is a labelling function. 
% \begin{itemize}
% 	\item $S$ is a countable, non-empty set of states
% 	\item $Act$ is a set of actions
% 	\item $\mathbf{P}: S \times Act \times S \to [0,1]$ is a transition probability function such that for all states $s \in S$ and actions $\alpha \in Act: \sum_{s' \in S} \mathbf{P}(s,\alpha,s') \in \{0,1\}$
% 	\item $p_i: S \to [0,1]$ is the initial state distribution, such that $\sum_{s \in S} p_i(s) = 1$
% 	\item $AP$ is a set of atomic propositions, and
% 	\item $L: S \to 2^{AP}$ is a labelling function. 
% \end{itemize}
% An action $\alpha \in Act$ is enabled in a state $s$ if, and only if, $\sum_{s' \in S} \mathbf{P}(s,\alpha,s') = 1$, and $Act(s)$ denotes the set of actions enabled in $s$.
The set of actions available in state $s\in S$ is denoted by $Act(s)$.
\end{defi}
% \begin{defi}
% Paths in DTMCs and MDPs

% An \textit{(infinite) path} $\pi$ in a discrete-time Markov chain $\mathcal{M} = (S, \mathbf{P}, p_i, AP, L)$ is defined as an infinite sequence of states $\pi = s_0s_1s_2...$ such that $\forall i \geq 0: \mathbf{P}(s_i, s_{i+1}) > 0$.

% % A \textit{finite path} $\rho$ in the discrete-time Markov chain $\mathcal{M}$ is defined as the prefix of an infinite path $\pi$, $\rho = s_0 \to s_1 \to ... \to s_n$ for $n > 0$.

% An \textit{(infinite) path} $\pi$ in a Markov decision process $\mathcal{M} = (S, Act, \mathbf{P}, p_i, AP, L)$ is defined as an infinite sequence of states and actions $\pi = s_0a_0s_1a_1s_2...$ (also written as $\pi = s_0 \xrightarrow{a_0} s_1 \xrightarrow{a_1} s_2 ...$) such that $\forall i \geq 0: \mathbf{P}(s_i, a_i, s_{i+1}) > 0$.
% % A \textit{finite path} $\rho$ in the Markov decision process $\mathcal{M}$ is defined as the prefix of an infinite path $\pi$, $\rho = s_0 \xrightarrow{a_0} s_1 \xrightarrow{a_1} ... \xrightarrow{a_{n-1}} s_n$ for $n > 0$.
% % In both Markov chains and decision processes, the set of all infinite paths is denoted by $Paths(\mathcal{M})$ while the set of finite paths is denoted by $Paths_{fin}(\mathcal{M})$
% \end{defi}
\begin{defi}[Path \& Policy]
A \textit{finite path} of an MDP is a finite sequence of states $s_0s_1\ldots s_n$ such that the transition probability from $s_{i}$ to $s_{i+1}$ is non-zero under some action in $Act(s_{i})$ for all $i\in \{0,\ldots,n-1\}$.  The set of all finite paths are denote by $S^*$.
A \textit{policy} for an MDP 
% $\mathcal{M} = (S, Act, \mathbf{P}, s_0, AP, L)$ 
is a function $\pi: S^* \to Act$ that maps a finite path to an action such that $\pi(s_0s_1 \ldots s_n) \in Act(s_n)$. The set of all policies is denoted by $\Pi$.
% An adversary $\sigma$ is called memoryless if, and only if, for any $\pi_1 = s_0s_1...s_n$ and $\pi_2 = s'_0s'_1...s_n$ it is true that $\sigma(\pi_1) = \sigma(\pi_2) = \sigma(s_n)$.
\end{defi}

\section{Abstraction and Verification of the Human-Vehicle System}
\label{sec:verification}

To verify the human driver model under a specification $\varphi$, we first abstract it to a Markov Chain $\mathcal{M}_h$. We achieve this by discretizing the individual modules of the integrated human driver ACT-R model in \cite{salvucci_1} through the use of the vehicle model. We can then use off-the-shelf tools, e.g., PRISM 
\cite{prism}, to perform the verification of the abstracted model. Below, for the purpose of clarify of presentation, we detail the abstraction procedure for a two-lane highway scenario, but we emphasize that the method extends trivially to $n$ lanes.

\subsection{Control Module}
\label{sec:control}

The control module of ACT-R is fully deterministic and can be divided into lateral (i.e. steering) and longitudinal (i.e. acceleration) control. The lateral control is determined by the existence of two artifacts that the driver obtains using low-level perception cues: the near and far points. In each ACT-R cycle, the model uses perception to determine the difference in visual angles $\Delta \theta_\near$ and $\Delta \theta_\far$ and the difference control law for the steering angle $\rho_h$ is:
\begin{equation}
\label{eq:control_steer}
	\Delta \rho_h = k_\far \Delta \theta_\far + k_\near \Delta \theta_\near + k_I \min{(\theta_\near, \theta_{\max})} \Delta t,
\end{equation}
where $k_\far$, $k_\near$ and $k_{I}$ are proportional control gains, and $\theta_{\max}$ is the maximum steering angle \cite{salvucci_1}. The process for the longitudinal control is similar. In each ACT-R cycle, the model starts by encoding the position of the lead vehicle and calculating the time headway to it, as well as the difference between this and the previous cycle, $\Delta t^\hw_\car$. The difference control law for the acceleration $a_h$ can then be written as:
\begin{equation}
\label{eq:control_acc}
	\Delta a_h = k_\car \Delta t^\hw_\car + k_\follow (t^\hw_\car - t^\hw_\follow)\Delta t,
\end{equation}
where $k_\car$ and $k_\follow$ are proportional gains of the control, and $t^\hw_\follow$ is the threshold time headway for following a vehicle \cite{salvucci_1}. To initiate a lane change, the driver begins following the near and far points of the destination lane instead of the current one \cite{older_3}.

% The most direct approach to abstracting this module, widely seen in the literature for small scenarios (e.g. \cite{grid_planning_1, grid_planning_2, grid_planning_3}), is to represent the road as an $N\times M$ grid, in which a tuple $(x,y) \in \{0,\ldots ,N-1\}\times\{0,\ldots ,M-1\}$ describes the position of the ego-vehicle, and all other state variables are integer versions of their continuous model representation. The error associated with the use of the grid as a way to discretize space can be reduced with an increase of $N$ and $M$ for a representation of the same road segment (i.e. increase in resolution). However, this incurs in the problem of state explosion: as the variable ranges increase, the number of states in the system grows exponentially and the model checking becomes intractable.

The most direct approach to abstracting this module, widely seen in the literature for small scenarios (e.g. \cite{highway_pred, grid_planning_1, grid_planning_2, grid_planning_3}) is to represent the road as a grid with the position of the ego-vehicle being a cell in the grid. The error associated with this method of discretizing space can be reduced by decreasing the cell area, i.e., increase in resolution. However, this incurs in the problem of state explosion: as the resolution increases, the number of states in the system grows exponentially and the verification becomes intractable.

In this work, we take a different approach and focus on reducing the dimensionality of the problem into a less error-prone space. We project the human-vehicle system state $(x, y, v, \psi, a, \rho, t) \in \mathbb{R}^7$ to $\mathbf{x} = (x, v, \lambda, a, t) \in \mathbb{R}^4\times \{0, 1\}$, where $x$ is bounded to a finite length of the road given by the scenario $\mathcal{S}$, and $\lambda \in \{0, 1\}$ represents the index of the lane (left or right).  A time discretization is induced by $\Delta t$ for all the continuous variables. Note that $t$ is included in $\mathbf{x}$ to enable the tracking of the state of the other vehicle, whose motion is assumed to be known (see Sec. \ref{sec:kinematics}). 
We further reduce the representation by compressing the lane change maneuver into a single transition, as described below.
% When a lane change is decided, the controls and state of the vehicle are given by \eqref{eq:car-kinematics}-\eqref{eq:control_acc}. We declare the maneuver is complete when the vehicle has merged to the center of the final lane.

% \fge{reduce dim of state and compress the lane change into a single transition}
% The acceleration and velocity of the ego-vehicle are obtained from equations \eqref{eq:car-kinematics} and \eqref{eq:control_acc}, with time discretized with a fixed $\Delta t$ 
The evolution of the compressed model is as follows. When the vehicle is following its current lane, $\lambda$ remains the same, and $x$ and $v$ are given by \eqref{eq:car-kinematics} ($y$ is the center of lane $\lambda$ and $\psi=0$) with the control input $\Delta \rho_h$ being zero and $\Delta a_h$ given by \eqref{eq:control_acc}.
When a lane change is decided, the controls and state of the vehicle are given by \eqref{eq:car-kinematics}-\eqref{eq:control_acc}.  We declare the maneuver is complete when the vehicle has merged to the center of the final lane, updating $\lambda$.  During the maneuver, we monitor the change in the truth values of the atomic propositions in addition to possible collisions.  Then, we discard the maneuver trajectory and record only the two states, at which the lane-change maneuver starts and ends, and label the latter state with the propositions of the maneuver.  
These values can be pre-computed, stored in lookup tables, and used for deterministic transitions between the control and the next ACT-R step, producing significantly smaller models.

\subsection{Decision Making and Monitoring}
\label{sec:ACTR-decision}

The decision making process to move from the right to the left lane consists of localizing the lead vehicle in the right lane and deciding whether or not to change lanes based on the time headway, $t^\hw_\car$. The lower this time headway, the more likely a driver is to perform the manoeuvre \cite{thw}.
% Decision making in such a case follows from a stochastic reasoning \ml{is the following sentence really needed?  Keep in mind, you should present the framework in a general way.} which can also be simulated and incorporated in the model using look-up tables.
Let $d$ to be the distance between the two vehicles.
% and $v$ the speed of the ego-vehicle. 
We represent the probability of the driver performing a lane change to the left lane with an exponentially decreasing function (as in \cite{time_headway_exponential, thw_exponential}):
% \ml{justify why this is a good representation ... also say the framework is general and other distributions can be use}:
\begin{equation}
\label{eq:left_to_right}
	P_{l_c}(t^\hw_\car \cond \lambda = 0) = e^{-\alpha t^\hw_\car},
\end{equation}
where $\alpha$ is a parameter of the decision making. 
% While this exponentially decreasing distribution is assumed in this case, the framework is general and other distributions can be used for the same purpose.

A similar approach can be applied for a driver in the left lane overtaking a vehicle behind it in the right lane, except in this case the opposite effect occurs in the decision making. In such a case, the probability of changing lane can be modelled as a normalized logarithmic function over the distance between the vehicles:
\begin{equation}
\label{eq:right_to_left}
	P_{l_c}(d,v \cond \lambda = 1) = \frac{\log(\beta d + 1)}{\log(\beta d_{\max} + 1)},
\end{equation}
where $d_{\max}$ is the maximum length considered in the scenario, and $\beta$ is a parameter of the decision making.
% \ml{what's $d$? distance to the lead vehicle?}. 
% Similarly to the previous case, in this case other distributions can be used instead of the normalized logarithmic function. 
It should be noted that the values of $\alpha$ and $\beta$ could be estimated from real data for a population of drivers \cite{time_headway_exponential, thw_exponential}.

% For the purposes of analysis, three classes of drivers are considered: Aggressive, Average and Cautious drivers (with different values for the parameters $\alpha$ and $\beta$). 

% In Figure~\ref{fig:dm_curves}, the functions are presented for the three profiles of drivers assumed in this project, for the lane changes originating in the right (Figure~\ref{fig:dm_curve_right}) and left lane (Figure~\ref{fig:dm_curve_left}).

% \begin{figure}
% \centering
% \begin{subfigure}{1\textwidth}
% 	\centering
% 	\includegraphics[width=0.85\textwidth]{dm_curve.pdf}
% 	\subcaption{Aggressive ($\alpha = 1$), Average ($\alpha = 0.6$) and Cautious ($\alpha = 0.4$) drivers.}
% 	\label{fig:dm_curve_right}
% \end{subfigure}\\
% \begin{subfigure}{1\textwidth}
% 	\centering
% 	\includegraphics[width=0.85\textwidth]{dm_curve_2.pdf}
% 	\subcaption{Aggressive ($\beta = 1000$), Average ($\beta = 0.5$) and Cautious ($\beta = 0.01$) drivers.}
% 	\label{fig:dm_curve_left}
% \end{subfigure}\vspace{1em}
% \caption{$\text{P[}l_c = true\text{]}$ as a function of $t^\hw_\car$ and $d$ for vehicles originating from the right lane (a) and left lane (b).}
% \label{fig:dm_curves}
% \end{figure}

So far, this version of decision making is not influenced by the monitoring module at all, and it relies on the measurements of the values of $t^\hw_\car$ and $d$ by the human. It is unrealistic to assume that the human's measurements are perfect.  In order to reflect uncertainty in these values, stochastic noise is added to the measurement of $d$ (as this is what human drivers have to instinctively measure through perception).  The noise is considered to be normally distributed $w \sim \mathcal{N}(0, \sigma)$. For an integral resolution parameter, $\delta \in [0, d]$, and $L$ as the number of discrete steps for $x$, we can define: 
% \fge{delta in range 0 to d, it defines the resolution of the integral; talk about L as the number of discrete steps}:
\begingroup\makeatletter\def\f@size{10}\check@mathfonts
\def\maketag@@@#1{\hbox{\m@th\large\normalfont#1}}
\begin{equation}
\label{eq:monitoring}
\begin{aligned}
	& P'_{l_c}(d,v) = \sum_{i = -L}^{L} P_{l_c}(d + i,v) \int_{d + i - \delta/2}^{d + i + \delta/2} w(z) dz.
\end{aligned}
\end{equation}
\endgroup
% \ml{$L$ or $x_{\max}$?  Make sure the integral limits are correct! $1/2$ or $d/2$?}
% defined as:
% \begin{equation}
% 	N(d) = \int_{-\infty}^d n(t) dt
% \end{equation}

Similarly to the control module, the values of $P'_{l_c}$ can be pre-computed and stored in a table to be used in stochastic transitions to the ACT-R control step of the following cycle.

% From this, a look-up table can be generated which, for different driver types, yields the probability of lane changing for a certain distance to the lead car and velocity.
% For the 3 driver profiles mentioned, considering $d \in \{1,...,80\}$ ($\max_d = 80$) and $v \in \{15,...,34\}$, the table generated for the vehicle in the right lane has $4,800$ rows. Under the same conditions, the look-up table for the vehicle in the left obtained has $240$ rows (the difference is explained by the fact that the velocity does not influence this table). Thus, the unified decision making table has $5,040$ rows.

% \section{UNIFIED TWO-MODULE MODEL}
\subsection{Markov Chain Abstraction}

We now define a finite MC $\mathcal{M}_h = (S, \mathbf{P}, s_0, AP, \iota)$ that 
% Using the abstractions generated, the final Markov Chain model 
unifies both modules using
the discretization described above and
a variable $\mu \in \{1,2\}$, where $\mu = 1$ corresponds to the control step and $\mu = 2$ to the decision making stage.  
We define a state $s \in S$ of $\mathcal{M}_h$ to be a tuple $s = (\mu, x, \lambda, a, v, t)$.  For a given scenario $\mathcal{S} = (\lambda_0, x_0, v_0, \mathbf{x}^{ov})$, where $\mathbf{x}^{ov}$ is the state of the other vehicle, the state space $S$ is automatically generated.
% are generated automatically for a given scenario $\mathcal{S} = (\lambda_0, x_0, v_0, x^{ov})$,
% \ml{$\mathcal{S}$ should also include the initial lane or $y_0$!},
% \ml{should also consider the initial state of the car: $\mathcal{S} = (x_0, v_0, x^\ev_0, v^\ev_0)$}
The transition probabilities for all $s, s' \in S$ are given by:
\begin{equation*}
    \mathbf{P}(s,s') = 
    \begin{cases} 
        1 & \text{if } \mu_s = 1 \wedge s' = \text{\textsc{Control}}(s), \\
        \textsc{DMM}(s, s') & \text{if } \mu_s = 2, \\
        0 & \text{otherwise,} 
    \end{cases}
\end{equation*}

\noindent
where \textsc{Control} is the lookup table for the control step described in Sec \ref{sec:control} and \textsc{DMM} is the probability table for the decision making and monitoring stage described in Sec \ref{sec:ACTR-decision}.
% and the function \textsc{PDist}, which represents whether the distance between the ego-vehicle and the other vehicle is positive or not (that is, is the other vehicle in front or behind the ego-vehicle), is defined as\footnote{\textsc{sgn} is the commonly defined sign function}:
% \begin{equation}
% \textsc{PDist}(x, t) = \frac{1}{2} \cdot (\textsc{sgn}[x - x^{ov}(t)] + 1) \in \{0,1\}
% \end{equation}
The set $AP$ and labeling function $\iota$ are naturally mapped according to the tuple elements of each state $s$.
It is important to note that the generated model is symbolic in nature, adding to the flexibility of the framework. Furthermore, it is worth noting that $\mathcal{M}_h$ captures in a one-to-one mapping all the possible outcomes of the continuous integrated driver model, under the assumptions of the distributions given by \eqref{eq:left_to_right}, \eqref{eq:right_to_left} and \eqref{eq:monitoring}.

% \ml{explain this equation... what is this distribution?  what does it describe?}
% \fge{Note that this is a symbolic representation}

\subsection{Verification of the Human-Vehicle System}

Given the model $\mathcal{M}_h$, we are interested in computing the probability of satisfying a property $\varphi$ for a given scenario $\mathcal{S}$, i.e, Problem \ref{problem:1}. This probability is defined as:
\begin{equation}
    \mathbb{P}^{\mathcal{S}} (\varphi) = \text{Pr}(s_0 \models \varphi),
\end{equation}
that is, the probability of $\varphi$ holding in $\mathcal{M}_h$ from an initial state $s_0$. This problem has been extensively studied in the literature \cite{bk08}, and many linear programming based solutions for it exist using off-the-shelf tools, e.g. PRISM \cite{prism}, hence solving Problem \ref{problem:1}.

%%%%%%%%%%%%%%%%%%%%%%%%%%%%%%%%%%%%%%%%%%%%
%%%%%%%%%%%%%%%%%%%%%%%%%%%%%%%%%%%%%%%%%%%%
\section{Synthesis Framework for the ADAS}
\label{sec:synthesis}
In this section, we focus on the design of the ADAS and its representation as an MDP.

% The design of a correct-by-construction assistance system corresponds to determining which actions are available to the system at each state (the model becomes an MDP), and then performing synthesis to obtain the policy that maximizes/minimizes a given specification. The actions must be realistic in nature, otherwise the obtained assistance system would prove to be useless in a real-world scenario. It is assumed that the assistance system can not change the decision making (as it is a human cognitive process), but it can influence it to a certain degree through suggestion \cite{driver_behavior}. In the interest of safety and efficiency, incremental control-based options are also explored, both at the level of acceleration, as well as at the level of steering control, as summarized in Fig.~\ref{fig:final_system}.

\subsection{Passive Suggestions}
\label{sec:passive_suggestions}

At the decision making level, the human driver model in \cite{salvucci_1} has two options: it can either change lane or continue in the current lane.  These options can be influenced using suggestions (e.g. through visual or auditive cues)  \cite{driver_behavior}.  If a driver can be influenced to make a conscious decision to decelerate (e.g. through the suggestions of the ADAS), then there is an argument for including this action in the decision making.  Thus, we consider a 3-option ADAS with the following set of action suggestions:
\begin{equation*}
    Act = \{\act_\text{cl}, \, \act_\text{con},  \, \act_\text{dec}\},
\end{equation*}
where 
$\act_\text{cl}$, $\act_\text{con}$, and $\act_\text{dec}$ represent ``\textit{change lane}'', ``\textit{continue driving in this lane}'', and ``\textit{decelerate}'' respectively.  
% Therefore, the set of actions can be represented by
% % an action $\omega$ is defined as:
% \begin{equation}
% 	Act := \{(l_c), \big(\neg l_c \wedge (a_{s'} = a_{s}) \big), (\neg l_c \wedge a_{s'} = a_d)\}
% \end{equation}
% where $a_{s'}$ is the acceleration in the next state, $a_{s}$ is the acceleration of the vehicle in the current state and 
We assume that the human applies a constant deceleration value $a_d$ when the deceleration decision is made, i.e., $a_h = a_d$.  Then, we can abstract this human-vehicle-ADAS system as an MDP in a similar fashion to the MC abstraction above.  Note that the MDP includes additional states that correspond to the decision (action) of deceleration.  These states can be computed using the same procedure in Sec. \ref{sec:control} and the use of $a_h = a_d$.  The set of actions of the MDP is $Act$.

The transition probabilities of the MDP depend on how compliant the drivers are with the suggestions.
For the case that they are fully compliant, the decision making at each step can be replaced by all the possible actions in $Act$, obtaining an MDP with three deterministic transitions at this level.  However, full compliancy at all times is not realistic by any means.  To capture all possibilities, we define $\gamma \in [0,1]$ to be the responsiveness level of a driver to the suggestions given by the ADAS, where values 0 and 1 correspond to fully adamant and fully compliant driver, respectively.  Building on the framework in \cite{salvucci_1} and Sec. \ref{sec:ACTR-decision}, let $p$ be the probability that a driver decides to change lane at state $s$, i.e., probability of deciding to continue in the current lane is $1-p$.  
Furthermore, 
denote by $s'_i$, the successor of state $s$ if the vehicle performs action $i$. 
% Then, 
% denote by $s'_\text{lc}$, $s'_\text{con}$, and $s'_\text{dec}$, the MDP states that correspond to the vehicle changing lane,  continuing driving, and decelerating at state $s$.  Also, let $p$ the probability that the driver decides to change lane in state $s$.  
Then, the transition probabilities of the MDP from the states $s \in S$ that correspond to ACT-R decision making, i.e., $\mu_s=2$, are given by:
\begin{equation*}
    \footnotesize
    \mathbf{P}(s,\act, s') =
    \begin{cases} 
        \gamma + (1-\gamma)p & \text{if } \act=\act_\text{lc} \wedge s'= s'_\text{lc}, \\
        (1-\gamma)(1-p) & \text{if } \act=\act_\text{lc} \wedge s'= s'_\text{con}, \\
        % 0 & \text{if } \act=\act_\text{lc} \wedge s'= s'_\text{dec}, \\
        \gamma & \text{if } \act=\act_\text{dec} \wedge s'= s'_\text{dec}, \\
        (1-\gamma)p & \text{if } \act=\act_\text{dec} \wedge s'= s'_\text{lc}, \\
        (1-\gamma)(1-p) & \text{if } \act=\act_\text{dec} \wedge s'= s'_\text{con}, \\
        \gamma + (1-\gamma)(1-p) & \text{if } \act=\act_\text{con} \wedge s'= s'_\text{con}, \\
        (1-\gamma)p & \text{if } \act=\act_\text{con} \wedge s'= s'_\text{lc}, \\
        % 0 & \text{if } \act=\act_\text{con} \wedge s'= s'_\text{dec}, \\
        0 & \text{otherwise} 
    \end{cases}
\end{equation*}

\noindent
Note that, since $\gamma, p \in [0,1]$, the transitions are guaranteed to sum up to one under each action. 
% The decision making with fully compliant drivers corresponds to the case where $\gamma = 1$.

\subsection{Active Control}

\paragraph*{Active Acceleration Control}
% Considering the assumptions previously described, 
Active acceleration control by the ADAS is an incremental addition to the acceleration values applied by the human in the control module, i.e., $a = a_h + a_\text{ADAS}$.  Let the acceleration of the vehicle 
% given by $a \in \{a^{\min},...,a^{\max}\}$. 
bounded by $a \in [a^{\min},a^{\max}]$.
In this module, a value $a_\text{ADAS} \in \{a_\text{ADAS}^{\min}, \ldots , a_\text{ADAS}^{\max}\}$ is considered such that:
\begin{equation}
\label{eq:delta_as_rest}
	a_\text{ADAS}^{\min} > a^{\min} \quad \text{ and } \quad a_\text{ADAS}^{\max} < a^{\max}.
\end{equation}
Hence, the final acceleration applied to the vehicle becomes
\begin{equation}
	a = \max (\min (a_h + a_\text{ADAS}, a^{\max}), a^{\min}).
\end{equation}
The restriction to the values of $a_\text{ADAS}$ presented in \eqref{eq:delta_as_rest} allows the system to be incremental instead of enforcing the specific values chosen by the ADAS, i.e., it is corrective instead of assertive. 
% This is important to avoid policies which sharply contrast in terms of the values the human.
% , e.g. a strategy which at a certain time chooses an acceleration of $3$ and in the next time step chooses one of $-2$ (not allowed for a small enough range of $\Delta a_{as}$ and according to the linear control abstraction obtained in the previous section). 
% For implementation purposes, it is assumed that $\Delta a_{as} \in \{-1,0,1\}$.

% The implementation of such a system consists in replacing the existing linear control at each step of the control module by the resulting accelerations of applying all the possible values of $\Delta a_{as}$ to the acceleration decided by the human control module, obtaining an MDP with at most three actions at the acceleration control level (and at least two).

\begin{figure}
    \centering
    \includegraphics[width=0.47\textwidth]{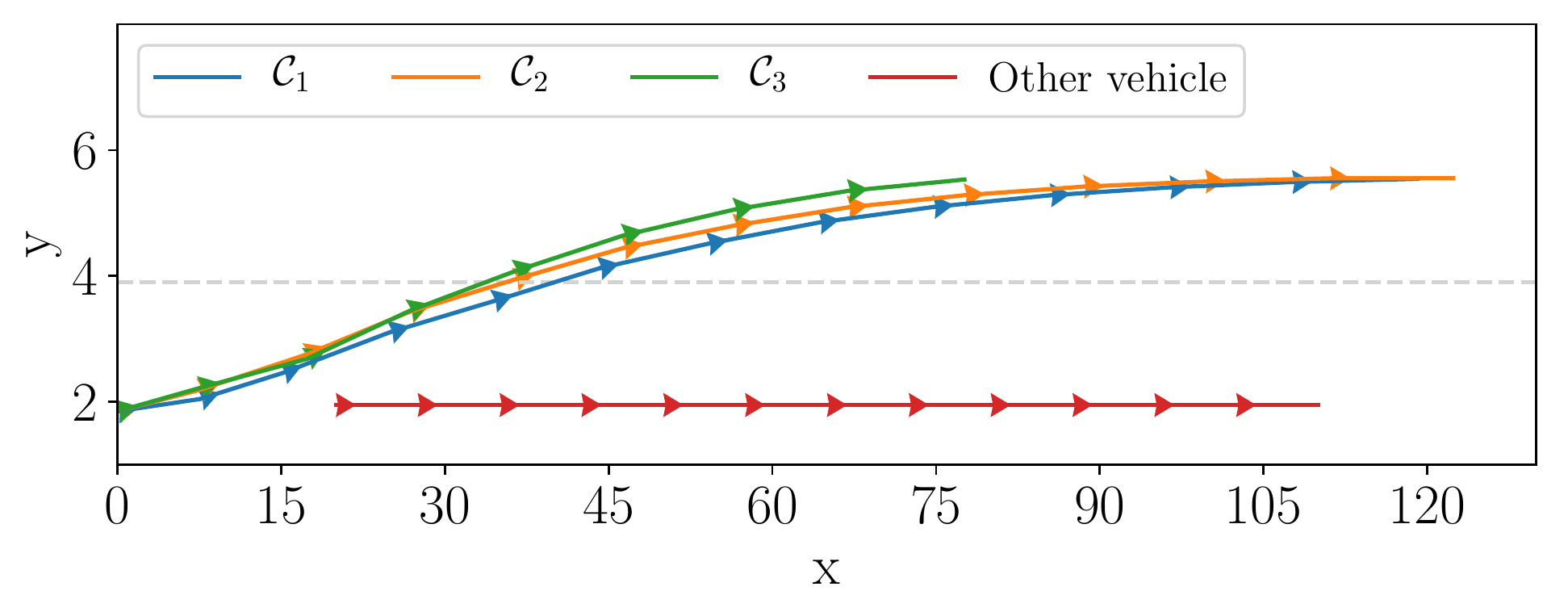}
    \caption{Example of the simulation of a lane change ($x$ and $y$ in meters) for 3 different sets of parameters $\mathcal{C}_i = (k_\far^\text{ADAS}, k_\near^\text{ADAS}, k_I^\text{ADAS})$ for the steering control law, with $\mathcal{C}_1 = (15, 3, 5)$, $\mathcal{C}_2 = (17, 3, 6)$ and $\mathcal{C}_3 = (14.5, 3, 7)$.}
    \label{fig:lane_change_ex}
\end{figure}

\paragraph*{Active Steering Control}
The human driver model in ACT-R uses the control law in \eqref{eq:control_steer} for the steering angle $\rho$ for given $k_\far, k_\near$ and $k_I$ \cite{salvucci_1}. 
We design the active steering control of the ADAS ($\rho_\text{ADAS}$) to be given by the same control law with different sets of gains.
% By changing the values of these constants, different control laws are obtained, which introduce different accelerations and velocities at each time step, mimicking the behavior of the incremental control previously assumed (i.e. within such a policy, the difference in acceleration and velocity introduced is the difference between these values for the two control laws at each time step). 
Thus, actions in this part of the assistance system at the model level correspond to different sets of $\mathcal{C}_i = (k_\far^\text{ADAS}, k_\near^\text{ADAS},k_I^\text{ADAS})_i$ available to the ADAS.  The resulting steering angle applied to vehicle $\rho = \rho_h + \rho_\text{ADAS}$ essentially becomes the control law in \eqref{eq:control_steer}, where the gains are the sum of gains for the human and ADAS, i.e., incremental (corrective) control, as exemplified in Fig.~\ref{fig:lane_change_ex}.
% For implementation purposes, three distinct sets of parameters were considered as possible actions, $(k_\far, k_\near,k_I) \in \{(15,3,5), (17,3,6), (14.5,3,7)\}$.

We augment the MDP obtained in Sec.~\ref{sec:passive_suggestions} by adding the deterministic actions described to the control stage of the model. That is the transition probabilities of the MDP for the states $s \in S$ that correspond to ACT-R control, i.e., $\mu_s=1$, are given by:
\begin{equation*}
    \mathbf{P}(s,\act,s') = 
    \begin{cases} 
        1 & \text{if } s' = \text{\textsc{ActControl}}(s, \act), \\
        0 & \text{otherwise,} 
    \end{cases}
\end{equation*}
where \textsc{ActControl} is the state-action lookup table for the active control step described above.
% by \textit{active acceleration} and \textit{steering control}.

% \ml{how is the MDP constructed?}

% \begin{figure}[h]
%     \centering
%     \includegraphics[width=0.9\textwidth]{steering_control_ex.pdf}
%     \caption{Example of the simulation of the lane change for $o_{lane} = 1$ (right lane), $d = 20m$, $v = 15m/s$ and $v_1 = 15m/s$ (the legend of each path corresponds to the situation with parameters $k_\far, k_\near,k_I$, respectively).}
%     \label{fig:steering_control_ex}
% \end{figure}

% Using these values, new look-up tables can be obtained with the probability of crashing, $\Delta x$ and $\Delta T$ incurred, and the final velocity of the ego-vehicle for each origin lane, distance to the other vehicle, velocities of both the vehicle in question and the other vehicle and the parameters of the control law (out of the three possibilities presented). Following the same calculations as shown in Section~\ref{sec:control}, the obtained look-up table contains $103,200$ rows.

% In terms of implementation, the MDP is obtained through simply reading the three possible actions for lane changing directly from the look-up table, similarly to the human driver model (the difference being the latter only has one option).

\subsection{Policy Synthesis}
\label{sec:policy-synthesis}
Recall that we are interested in designing an ADAS that optimizes the probability of satisfying a given property $\varphi$ for a scenario $\mathcal{S}$, i.e, Problem \ref{problem:2}.
The finite MDP constructed by adding the actions at the decision making and control levels, $\mathcal{M}_\text{ADAS}$, 
represents all the possible choices of the ADAS at every $\Delta t$ step of the driving scenario $\mathcal{S}$.  Therefore, the optimal ADAS problem is reduced to finding an optimal policy over $\mathcal{M}_\text{ADAS}$.   

\begin{figure*}[t]
\centering
\centering
\includegraphics[width=0.97\textwidth]{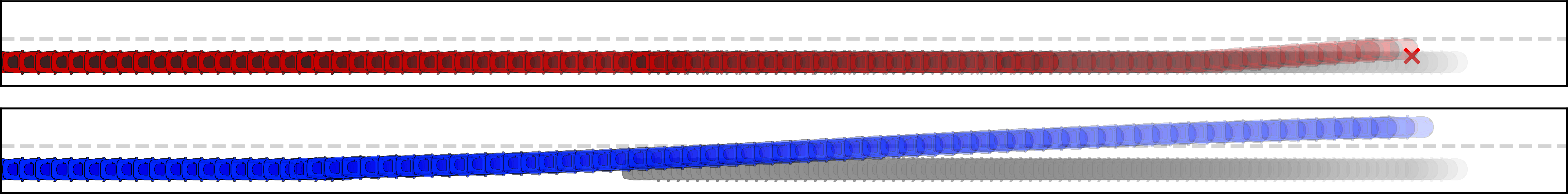}
\caption{
Example of a run for $\varphi_1$ with $\mathcal{S} = (0, 0 m,25 \frac{m}{s},50 m,15 \frac{m}{s})$. Top in red: human-vehicle system (no ADAS).  Bottom in blue: human-vehicle system with ADAS.  Gray: the other vehicle.  For readability purposes, the opacity of the cars decreases with time.  The red `x' marks a collision between the vehicles.
% Example of a run of the human-vehicle system model (top in red) and the system with the ADAS (bottom in blue) for $\mathcal{S} = (0, 0 m,25 \frac{m}{s},50 m,15 \frac{m}{s})$, with the other vehicle represented in gray; for readability purposes, the opacity of the cars decreases with time and the road segment shown is shortened to 120 meters; in the top plot, the red `x' marks a collision between the vehicles.
}
\label{fig:example_run}
\end{figure*}

\begin{figure*}[t]
\centering
\begin{subfigure}{0.35\textwidth}
  \centering
  \includegraphics[width=1\textwidth]{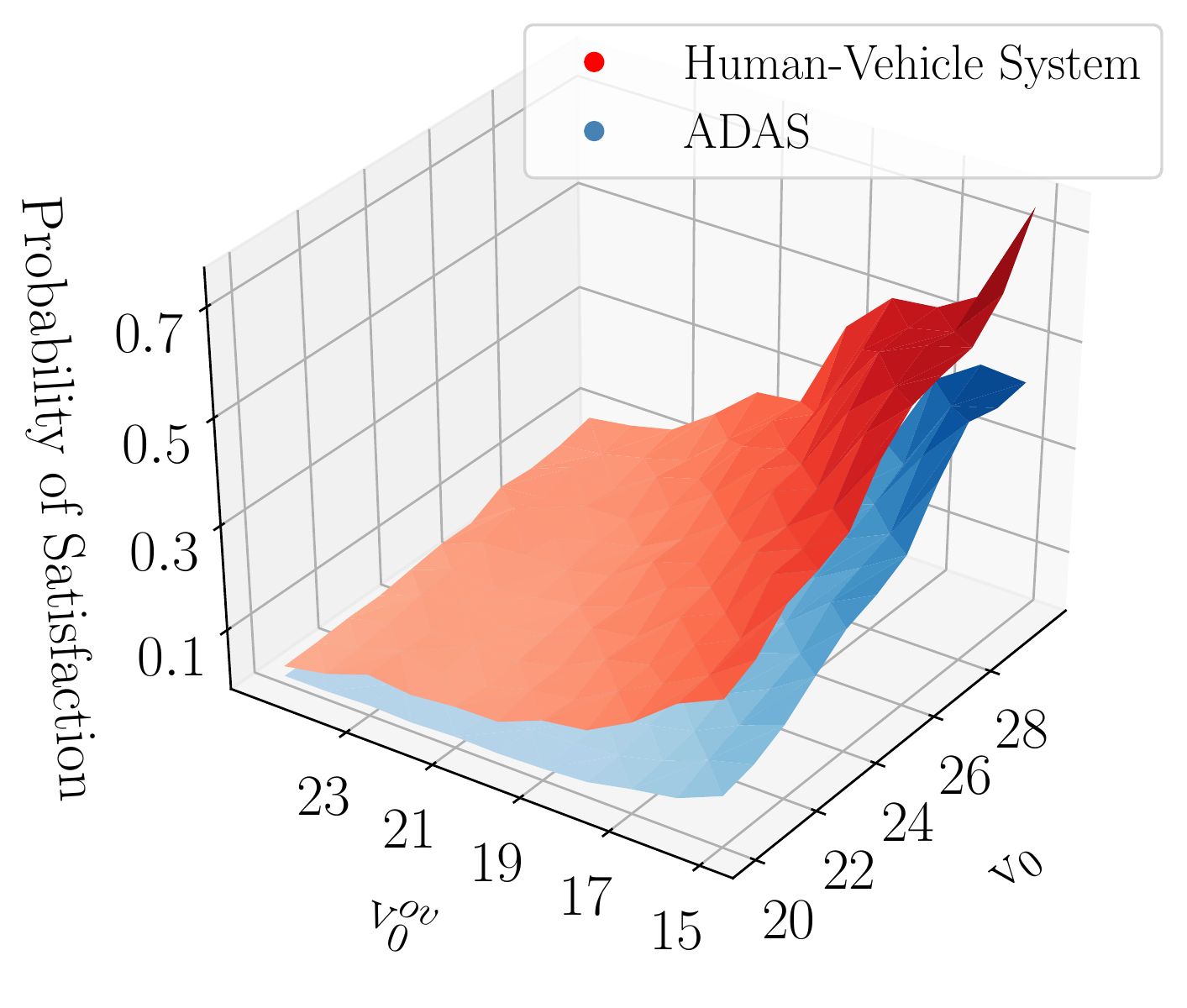}
  \subcaption{$\varphi_1$, $\bowtie = \min$}
  \label{fig:safety_3d}
\end{subfigure}
\begin{subfigure}{0.30\textwidth}
  \centering
  \includegraphics[width=1\textwidth]{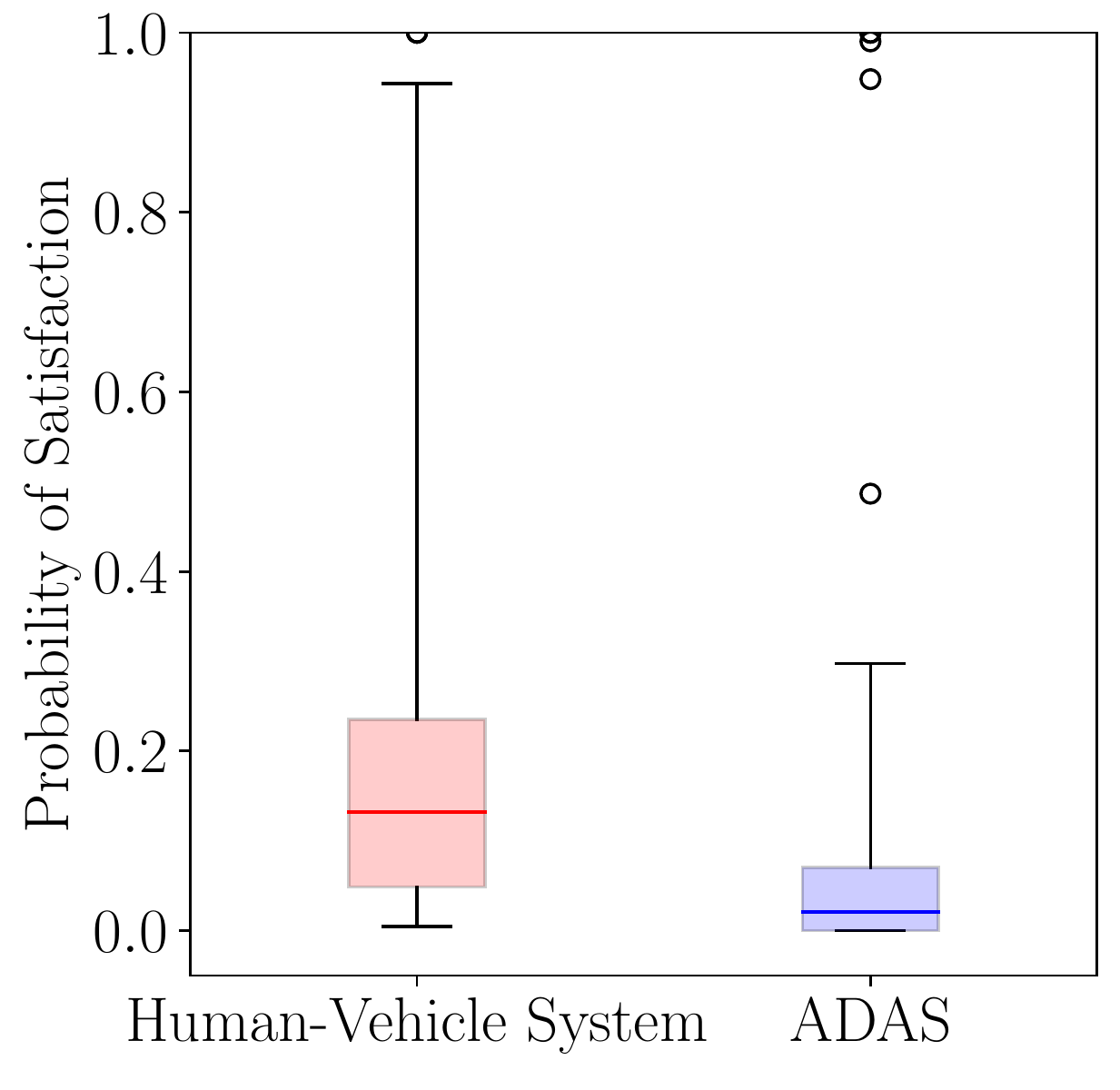}
  \subcaption{$\varphi_1$, $\bowtie = \min$}
  \label{fig:safety_box}
\end{subfigure}
\begin{subfigure}{0.30\textwidth}
  \centering
  \includegraphics[width=1\textwidth]{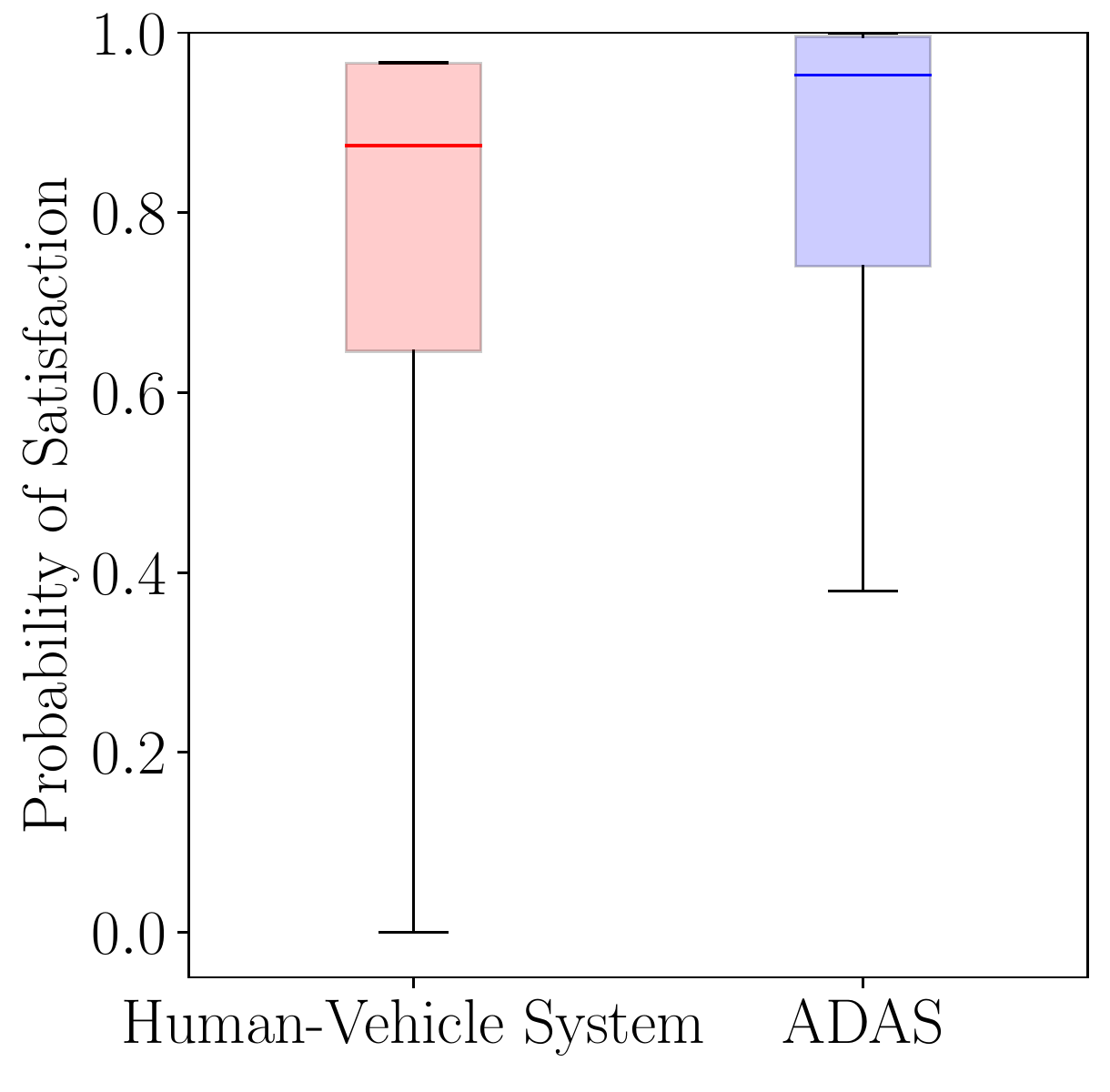}
  \subcaption{$\varphi_2$, $\bowtie = \max$}
  \label{fig:liveness_box}
\end{subfigure}
\caption{
Analysis of the probability of satisfaction of $\varphi_1$ and $\varphi_2$ in various conditions.  (a) varying scenarios $\mathcal{S} = (0, 0 m,v_0,50 m, v_0^{ov})$ with $v_0 \in \{20,...,30\} \frac{m}{s}$ and $v_0^{ov} \in \{15,...,20\} \frac{m}{s}$; (b) a randomly sampled population of 100 different scenarios $S$; and (c) a randomly sampled population of 100 different scenarios $S$ and $T=21s$.
% Variation of the satisfaction probabilities of $\varphi_1$ and $\varphi_2$ for the human-vehicle system (in red), $\mathbb{P}^{\mathcal{S}}(\varphi)$, and the model with the synthesized ADAS (in blue), $\mathbb{P}^{\mathcal{S}, \pi^*}_{\bowtie}(\varphi)$, in (a) the scenarios $\mathcal{S} = (0, 0 m,v_0,50 m, v_0^{ov})$ for $v_0 \in \{20,...,30\} \frac{m}{s}$ and $v_0^{ov} \in \{15,...,20\} \frac{m}{s}$; (b) a randomly sampled population of 100 different scenarios $S$; and (c) a randomly sampled population of 100 different scenarios $S$ and $T=21s$.
}
\label{fig:plot3}
\end{figure*}

The policy that maximally satisfies $\varphi$ is defined as:
\begin{equation}
	\pi^* \in {\arg\sup}_{\pi \in \Pi} \mathbb{P}^{\pi} (\varphi)
\end{equation}
and, respectively, $\arg\inf$ for minimally. 
% There are multiple references in the literature that cover methods to solve the optimization problem posed, in particular \cite{games, synthesis}. The probability of satisfying $\varphi$ under the optimal policy $\pi^*$ for a given scenario $\mathcal{S}$ is denoted $\mathbb{P}^{\mathcal{S}, \pi^*}(\varphi)$.
The computation algorithms for such policies are well-studied in the formal synthesis literature, and there exist many off-the-shelf tools, e.g., PRISM \cite{prism}, that solve this optimization problem efficiently.  In addition to the optimal policy $\pi^*$, these tools compute the probability of satisfying $\varphi$ under $\pi^*$, denoted by $\mathbb{P}^{\pi^*}(\varphi)$.

\section{Experimental Results}
\label{sec:expriments}

The proposed framework is implemented as an open source tool in Python using PRISM\footnote{\href{https://github.com/fgirbal/cbc\_adas}{Github repository: https://github.com/fgirbal/cbc\_adas}}. To illustrate its efficacy, we performed a series of case studies using various scenarios and specifications.  Due to space constraints, we can show only two of them here.  We refer the reader to \cite{thesis} for the full report on all the case studies.  

We considered a two-lane highway scenario with both the ego-vehicle and lead vehicle driving on the right lane on a road segment that is 500 meters long ($x_{\max} = 500m$).  The lead vehicle is assumed to be moving at a constant speed with $\mathbf{x}^{ov}(0) = (x_0^{ov}, v_0^{ov})$. The analysis below is performed based on the ACT-R parameters given in \cite{salvucci_1}.

\textit{Case Study 1:}
We are interested in minimizing ($\bowtie = \min$) the safety property of crashing, i.e, $$\varphi_1 = \Diamond \, \textsc{crash},$$ for initial conditions given by  $\mathcal{S} = (\lambda_0,x_0,v_0,x_0^{ov}, v_0^{ov})$. 
% $= (0 m,25 \frac{m}{s},50 m,15 \frac{m}{s})$.with no ADAS intervention, the verification framework generates $\mathbb{P}^S(\Diamond \, \textsc{crash}) = 0.489$.  This high probability of crashing is mainly due to the high initial speed of the ego-vehicle, leaving the human with a small room for mistake.  Fig. \ref{fig:example_run} captures two simulation runs, showing how a crash can occur and avoided. By adding ADAS to the ego-vehicle, however, this probability is reduced almost by half, i.e., $\mathbb{P}^{S,\pi^*}_{\min}(\Diamond \, \textsc{crash}) = 0.242$, showing its effectiveness. 
For $\mathcal{S} = (0, 0 m,25 \frac{m}{s},50 m,15 \frac{m}{s})$ with no ADAS intervention, the verification framework generates $\mathbb{P}^S(\varphi_1) = 0.489$.
% This high probability of crashing is mainly due to the high initial speed of the ego-vehicle, leaving the human with little room for mistakes. 
In this situation, the ego-vehicle is travelling at a high speed when compared to the lead vehicle, leaving the human with little room for mistakes. The constraints imposed by the human cognitive modeling, such as memory decay, distraction and limited motor performance, inevitably lead to a high probability of crashing.  By adding the ADAS to the ego-vehicle, however, this probability is reduced by more than half to $\mathbb{P}^{S,\pi^*}_{\min}(\varphi_1) = 0.242$, showing the effectiveness of the ADAS.  Fig.~\ref{fig:example_run} shows an example run for the human driver model (top), which results in a crash, and ADAS system (bottom), which avoids a crash by suggesting and actively contributing to the lane changing action early on.

% The results are presented in Fig.~\ref{fig:example_run} and \ref{fig:plot3} for both with and without ADAS. 
% Fig.~\ref{fig:example_run} shows an example run for the human driver model 
% and the full system in the same scenario, in which the human driver model observes a collision between the two vehicles while the full system with the ADAS avoids it by changing lanes earlier. It should be noted that, while this is an example run, the quantitative results of the properties are $\mathbb{P}^{\mathcal{S}}(\Diamond \: \textsc{crash})=0.489$ (without ADAS) and $\mathbb{P}^{\mathcal{S}, \pi^*}_{\min}(\Diamond\: \textsc{crash})=0.242$ (with ADAS). 

Fig.~\ref{fig:safety_3d} presents the variation of the probability of satisfaction of the safety specification with the change of $v_0$ and $v^{ov}_0$. As it can be observed, the introduction of the ADAS reduces the probability of crashing significantly in all the cases.
Fig.~\ref{fig:safety_box} shows boxplots for the same safety property in a randomly generated sample of 100 different scenarios, obtained by uniformly sampling over bounded intervals for each of the variables. In this case, it is also observed a decrease in the probability of satisfaction of $\varphi_1$, with the first, second and third quartiles in Fig.~\ref{fig:safety_box} being lower for the system with the ADAS than those for the human driver alone. 

\paragraph*{Case Study 2}
We are interested in maximizing ($\bowtie = \max$)  the liveness property of completing the road segment in under $T$ seconds, i.e, $$\varphi_2 = (\neg \textsc{crash}) \: \mathcal{U} \: \big( (x=x_{\max}) \wedge (t\leq T) \big),$$ for various sets of initial conditions given by  $\mathcal{S}$. % = (\lambda_0, x_0,v_0,x_0^{ov} v_0^{ov})$. 

Fig.~\ref{fig:liveness_box} shows boxplots of the probability of the liveness property for $T = 21s$ in a randomly generated sample of 100 different scenarios $\mathcal{S}$, obtained using the method previously described. In this situation, an increase in the probability of satisfaction of $\varphi_2$ is observed, with the first, second and third quartiles in Fig.~\ref{fig:liveness_box} being higher for the system with the ADAS than those for the human driver alone. 
This again illustrates the efficacy of the ADAS in terms of the satisfaction of the liveness property, i.e., the ADAS makes the system reach the end of the road safely and faster as required by $\varphi_2$.

\section{Final Remarks}
In this work, we proposed a framework for providing guarantees in ($i$) analyses of semi-autonomous driving scenarios and ($ii$) designing ADAS through the means of formal methods and modeling of the driver's cognitive process.  We achieved this by employing ACT-R to represent the human and a novel abstraction method that enables the representation of the infinite, continuous system of human-vehicle by a finite Markov model.
In the future, a data driven approach should be followed to validate the obtained results and evaluate the assumptions made about the drivers. A similar perspective can be taken for the design of the specifications, which could be learned in a closed loop fashion to minimize the difference between the full system with the ADAS and expert drivers. This is only possible due to the flexibility of the specifications allowed in the framework. Once the models are accurate according to the real world data, it is possible to deploy the obtained solutions.

\bibliographystyle{IEEEtran}
\bibliography{refs}

\end{document}